\documentclass{article}
\usepackage[nonatbib,final]{neurips_2022}
\usepackage[utf8]{inputenc}
\usepackage[T1]{fontenc}
\usepackage{cite}
\usepackage{amsmath,amssymb,amsfonts}
\usepackage{algorithmic}
\usepackage{graphicx}
\usepackage{textcomp}
\usepackage{xcolor}
\usepackage{tikz}
\usepackage{adjustbox}
\usepackage{placeins}
\usepackage{multirow}
\usepackage{array}
\usepackage{tabularx}
    \newcolumntype{L}{>{\raggedright\arraybackslash}X}
\usetikzlibrary{positioning, backgrounds, calc, arrows}
\def\BibTeX{{\rm B\kern-.05em{\sc i\kern-.025em b}\kern-.08em
    T\kern-.1667em\lower.7ex\hbox{E}\kern-.125emX}}
\graphicspath{ {result_images/} }
\usepackage{hyperref}       
\usepackage{url}            
\usepackage{subcaption}
\usetikzlibrary{shapes.geometric}
\usepackage{pgfplots}
\usepackage{bbm}
\usepackage{booktabs}
\usepackage{multirow}

\usepackage[utf8]{inputenc} 
\usepackage[T1]{fontenc}
\usepackage{booktabs}

\newcommand{\tsn}[1]{{\left\vert\kern-0.25ex\left\vert\kern-0.25ex\left\vert #1 
    \right\vert\kern-0.25ex\right\vert\kern-0.25ex\right\vert}}
\usepackage{xcolor}
\definecolor{darkred}{RGB}{150,0,0}
\definecolor{darkgreen}{RGB}{0,150,0}
\definecolor{darkblue}{RGB}{0,0,200}

\newcommand{\beq}{\begin{equation}}

\newcommand{\eeq}{\end{equation}}

\newcommand{\x}{\vct{x}}

\newcommand{\y}{\vct{y}}

\newcommand{\p}{{\vct{p}}}

\newcommand{\vct}[1]{\bm{#1}}

\def \endprf{\hfill {\vrule height6pt width6pt depth0pt}\medskip}

\makeatletter
\providecommand{\@noticestring}{}
\let\@notice\relax                 
\makeatother

\begin{document}

\title{Simulated Human Learning in a Dynamic, Partially-Observed, Time-Series Environment}
\author{%
   Jeffrey Jiang\textdagger\thanks{Department of Electrical and Computer Engineering, University of California, Los Angeles} \\
   \texttt{jimmery@ucla.edu} \\
   \And
   Kevin Hong\textdagger$^*$ \\
   \texttt{kevinhong1167@ucla.edu} \\
   \And
  Emily Kuczynski\textdagger$^*$ \\
  \texttt{ekuczynski@ucla.edu} \\
   \And
   Gregory Pottie$^*$ \\
   \texttt{pottie@ee.ucla.edu} \\
}
\def\thefootnote{\textdagger}\footnotetext{Equal contribution}\def\thefootnote{\arabic{footnote}}




\newcommand{\algor}{{{Simulated Education}}}
\newcommand{\alg}{{\textsc{SimEdu}}}

\newcommand{\mean}[1]{\mathbb{E}\left[ {#1} \right]}
\newcommand{\bern}[1]{\text{Bern}\left({#1} \right) }
\newcommand{\prob}[1]{\mathbb{P}\left[ {#1}\right]}
\newcommand{\vari}[1]{\textbf{var}\left[ {#1}\right]}
\newcommand{\covar}[1]{\textbf{cov}\left[ {#1}\right]}
\newcommand{\Reals}{\mathbb{R}}
\newcommand{\xvec}{\mathbf{x}}
\newcommand{\yvec}{\mathbf{y}}
\newcommand{\Amat}{\mathbf{A}}
\newcommand{\Xmat}{\mathbf{X}}
\newcommand{\eye}{\mathbf{I}}

\maketitle
\begin{abstract}
Human-machine interactions have become an increasingly important question as AI tools spread, incentivizing us to understand and adapt to the dynamics of humans. 
While intelligent tutoring systems (ITSs) can use information from past students to personalize instruction, each new student is unique. Moreover, the education problem is inherently difficult because the learning process is only partially observable. 
We therefore develop a dynamic, time-series environment to simulate a classroom setting, with student-teacher interventions—including tutoring sessions, lectures, and exams. In particular, we design the simulated environment to allow for varying levels of probing interventions that can gather more information. Then, we develop reinforcement learning ITSs that combine learning the individual state of students while pulling from population information through the use of probing interventions. These interventions can reduce the difficulty of student estimation, but also introduce a cost-benefit decision to find a balance between probing enough to get accurate estimates and probing so often that it becomes disruptive to the student. 
We compare the efficacy of standard RL algorithms with several greedy rules-based heuristic approaches to find that they provide different solutions, but with similar results. We also highlight the difficulty of the problem with increasing levels of hidden information, and the boost that we get if we allow for probing interventions. We show the flexibility of both heuristic and RL policies with regards to changing student population distributions, finding that both are flexible, but RL policies struggle to help harder classes. Finally, we test different course structures with non-probing policies and we find that our policies are able to boost the performance of quiz and midterm structures more than we can in a finals-only structure, highlighting the benefit of having additional information. 

\end{abstract}

\section{Introduction}

Human-machine interactions have become an increasingly important question as AI tools spread. Dealing with humans, however, introduces many challenges. First, people are necessarily dynamic, both in that a person's approach to interacting with tools may change quickly due to unobservable external factors and in that each human's response may have long-term, unobservable, or unknown dependencies. Secondly, there will almost never be enough data to train any model with any individual human, as the amount of experiences an individual human can have is limited. We turn to education as a prime example of a human-machine interaction that has long-standing research and interest, but provides a concrete example of the problems that we face. 

Artificial intelligence's role in education has become an increasingly important topic lately with the introduction of large language models (LLMs), which provide students with custom responses and artificial discussions about almost any searchable topic on the Internet. However, in this study, we seek to understand how to best create a long-term educational assistant for an individual, targeted toward assisting with classes. Intelligent Tutoring Systems (ITS) are computer-based educational tools that provide adaptive instruction to learners and are considered ``intelligent'' enough to substitute for human tutors. When considering ITS in the STEM domains, there is an inherent focus on model building and individual progress tracking \cite{ITS-review}. As an ITS tracks a student's progress, it will make interventions that will optimize the student's success in the course.

There are many points that make tracking human learning a difficult task for machines. In particular, education is a time-series problem with considerable hidden information. Hidden information occurs in many forms in the education environment, most clearly in student concept mastery, but also in terms of the possible responses that a student may have to a lecture and the effect of external factors. Each of these kinds of hidden information increases the difficulty of the problem, as we never have enough data from a single student to properly estimate everything, so we have to rely on a mixture of population knowledge and probing. Probing refers to the specific action by which the ITS either asks or quizzes student progress to update its estimate of the student's hidden concept mastery. 
The time-series nature of the problem adds to its difficulty, as if we save too much past information, the problem becomes intractable, while if we save too little, a past event may become a ``hidden external event'' from the perspective of the ITS. 

To gain better insight of how to best design an ITS, we breakdown many of the benefits and difficulties of the education problem and create a simulated environment. This simulated environment allows us to explore both realistic and idealistic scenarios with different techniques. 


In this chapter, we explore the options for creating an optimal ITS under many different simulated environment configurations. Among the questions we would like to answer are: 
\begin{enumerate}
    \item Can we create an ITS that can adapt to a student's individual learning characteristics? Struggling students should be recommended some remedial action, while the same recommendations should not apply to a student making good progress. 
    \item What is the value of probing? Would an ITS that cannot probe do significantly worse than one that can? And then, if the cost of probing increases, at what point would we no longer find it worthwhile to probe? 
    \item Can we design a course structure that is more robust to poor probing? That is, if our probing capabilities are limited, would a quiz structure be significantly better than a finals-only or midterm-final structure? 
\end{enumerate}

\section{Background}

\subsection{Motivation}
Recent years have enlightened many more of the general public of the current state and the future potential of artificial intelligence (AI) in our society. 
In light of this new frontier of AI, an important area of research is to understand how to interface the use of AI with humans.
Examples can be seen in all sorts of examples, such as self-driving cars, medical diagnosis, and education. 
AI has the potential to be extremely helpful to humans, but the blackbox nature of AI can also be hard to accept for many people and dangerous in extreme cases.
Thus, we would like to investigate ways to incorporate explainability and human interaction into the use of AI. Education can also be viewed as a proxy environment, providing some insight into the goals and problems of dealing with dynamic human interactions. 

Education is an interesting problem for artificial intelligence. On the positive side, typical educational programs are heavily structured and rules-based. This structure comes from many areas. There is a prerequisite concept structure for many classes, especially in STEM subjects.  Furthermore, we have a wealth of expert knowledge in how to teach specific courses, and a general idea for overall good teaching methodology. An example is that we know prerequisite concepts need to be at a certain level before any headway into new concepts can be made. 

However, education is challenging for artificial intelligence. Even outside the machine learning space, education is partially observed -- no instructor knows everything about a student. Partial observability can hinder the system's ability to adapt appropriately because it might not have access to all relevant contextual information, such as workload in other courses, extracurricular work and events, and family and personal relationships. This can make it difficult for the system to detect when a student is struggling, engaged, or in need of assistance. Without a complete picture of the learner's interactions, the system will miss opportunities to provide support, which can impact the learning process. 

Furthermore, education is a time-series problem. The state of a student is constantly changing with every point in time, especially if going to lecture, actively doing readings or assignments, and discussing with friends. As with many other instances of time-series data, time-series can pose several problems. First, we need to maintain an estimate of the student learning state, which depends, potentially, on points in time far in the past. It also depends on more parameters that we likely have no access to for the individual. Second, the true estimation of a student's state at a given point in time is effectively impossible without an extensive examination, detracting from time spent instructing and learning. 

Finally, explainability can be important, as students and teachers may be distrustful of computer recommendations. This can go in many ways, where extremes would be explaining to a struggling student why studying a specific concept is the most effective way to improve, but also in explaining to an excelling student that no further studying is necessary. Furthermore, we would like some method that is at least somewhat flexible in differing population distributions. These two points are closely related when it comes to methodology designs. 

\subsection{Prior Work}

Intelligent tutor systems (ITS) have been investigated for a long time and have been shown to outperform other computer aided instruction \cite{ITSvsCAI}. Much work has been contributed via the framework provided by ASSISTments \cite{online-edu-exp}. In particular, many experiments are run in the ASSISTments platform, and they include a long list of features that can quantify the students' skill level at any given point in time. In particular, this work has inspired and advanced the knowledge tracing domain significantly \cite{dkt, ktsurvey}. We use these features to create the simulated environment and acknowledge that we would like to do further analysis to see how much we can fit ASSISTment datasets within our own model. We use basic knowledge tracing in our population model, though we maintain explainability as our primary goal. 


The field of reinforcement learning (RL) has been a major driving force in the development of AI. 
RL has been used to solve many problems, such as playing games \cite{alphago, dqn, alphastar, alphazero}, robotics \cite{robotics}, and even medical applications \cite{MedicalIRL}. 
RL deals with an agent interacting with an environment and through the agent's interactions with the environment, learns how to best respond in the future. 
It combines the ability of online learning with the need to explore possible action consequences in the environment. 
An adjacent field, \textit{inverse reinforcement learning} (IRL) has also arisen in response to the popularity of RL, where we try to learn an optimal reward function that explains expert knowledge. 
While we don't directly reference IRL in this work, our use of prior knowledge can be easily extended to a discussion on IRL. 

A considerable amount of work has taken place for RL in the field of education \cite{edurl}. \cite{edudqn} tests the effectiveness of a Deep Q-Network (DQN), a Deep RL approach, on a simpler education environment, finding that it significantly outperforms heuristic methods. It shares further similarities in that it introduces a population model, using a standard single-step MLP, and a two-concept continuous-valued simulation course. We build upon this simulation by introducing deliberate hidden information, introducing probing and the trade-off of immediate time rewards, probes, and tutors, having a variety of concept mastery dynamics conditioned on student type, and a highly configurable simulation. 

\section{Problem Setting}
We view the education problem as a controls problem with unknown system dynamics. The autonomous intelligent tutoring system, a cognitive-dynamic system, is trying to give recommendations to steer the student (the system) in the correct direction. If all the mechanics of the system are deterministically known, course structure, student responses, etc., then we would have a traditional controls problem. 

This education problem is based on the interactions between students, teachers, and the overall structure of the course. Underlying everything, we assume that education, especially in STEM subjects comes with a hierarchical causal structure, \textit{e.g.} calculus builds on algebra, which builds on arithmetic. Whether a student learns a concept depends both on their mastery of prerequisite concepts and whether they have the motivation to spend time. 

Students may have the noble goal of achieving knowledge of the course material, but most of the time have the proxy goal of getting a good grade. 
As a student progresses through a course, she can take many actions and their state necessarily changes over time. However, each student's actions differ based on individual factors and are mostly unobserved. However, we consider that a student will consider a trade-off between extra performance in a course and doing other life activities. 

Similarly, professors generally want to educate students in their field, but also have other interests such as research, resulting in limited time. However, professors are typically unaware of the students' states. Therefore, traditionally courses have a hard time providing individualized support to students and require individual tutors.


Given this initial view, we introduce several major hurdles to directly understanding the creation of an ITS.  

\subsection{Hidden Information}
Education is a difficult task because of the various levels of hidden information. The control mechanics are not only stochastic, they are unknown and can be changing as a response to actions or due to completely unknown outside sources. At even the basic level, the two major hidden pieces of information at any point in time are the current concept state and the student's responses to any intervention that is taken. We will make restricting assumptions later to make the estimation problem feasible, but \textit{sensing} and \textit{probing} become the main ways to estimate the knowledge of the dynamics of the system. 

\subsection{Probing, Measurements, and Quantization} 

All measurements are imperfect and when it comes to human interactions, this is exacerbated by other psychological and social factors. In general, observations in the real world can only be measured up to some level of uncertainty, constrained by physical limits or just in measurement capacity of devices in general. In an ideal scenario, we would be able to directly know each student's mastery for all relevant concepts. However, this direct information cannot actually be obtained. Instead, this can only be measured through examinations, which are expensive in time for both students and instructors and are subject to measurement randomness, or through asking for student self-perception and historical performance, which can help us predict future performance. However, experiments we have run have shown that students often have no comparison metric to determine their own performance. Therefore, if possible, a quantitative measurement is preferred. 

On the other hand, we only have a limited number of control actions. There is a many-to-one relationship between possible states and the best possible action. Also, in education, intervention choices are less critical than in some other domains such as drone flight or medicine. These two properties combine to allow for some action ambiguity. As such, having a high precision state space is wasteful. It increases the necessary amount of parameters to draw precise boundaries, data to train such parameters, and difficulty in explanation. 

\subsection{Individual Dynamics}

Learning is inherently a dynamic process. 
Students build upon the information that they have already learned. 
Weak mastery of a prerequisite can hinder student developments on new dependent concepts. 
Furthermore, individual dynamics arise from both student differences and external factors. 
This can both be a reactionary response to something observable, e.g. an exam or homework, or unobservable, e.g. family trouble or financial issues.  

In a traditional Markov process, having full knowledge of the single time-step, $ \mathbb{P}\left[ s_{t} \vert s_{t-1} \right]$, fully characterizes the dynamics. 
Assuming this can be obtained, the main consideration lies in the decision of the state space, which can keep updated latent space information from past points in time. 
As students will respond to actions chosen by the ITS, we break down the state into the student's direct educational state and the student's observations of the ITS's actions as part of the student's observation space, notated as $ \mathbb{P}\left[ s_{t} \vert s_{t-1}, a_{t-1} \right]$. 
The Markov property is attractive, as it minimizes long-term graph connections, and allows prediction to occur independently of the past. 
With it, we can start helping a student no matter when they start using the ITS in the course (up to the warm-up period to filling out information about the student state $s_t$). 
Theoretically, with sufficiently large state spaces, the entire past can be represented in $s_t$. However, if this possible state space is too large, solving for the dynamics is intractable. 

Creating a relevant state space is thus nontrivial. In the education space, students may change dynamics due to unobserved events, possibly creating issues in the future. 
These could be due to events within education, such as poor teaching of a concept in a previous institution, or outside of education, such as prioritizing a part-time job. 
This can create situations where two students with the exact same history in observations in a course could suddenly diverge greatly. 
For any unobservable or unmodeled event, we require our dynamics to be flexible to sudden changes. This requires some consideration of the \textit{stability-plasticity dilemma} \cite{haykin}.

The other problem is that the dynamics themselves are an individual property. For every individual, $ \mathbb{P}\left[ s_{t} \vert s_{t-1}, a_{t-1} \right]$ will be different. 
Then, should the individual be included in the state space $s_t$? It is infeasible to do so, as we can never actually get enough data about the individual to accurately estimate the individual's transitions. 
Not only do individuals just not have the sheer number of interactions for a system to learn from, but also the actual instance of a transition is usually only relevant once. 
Therefore, we propose to estimate states of individuals assisted by what we have seen from the population.

\section{\alg{}}

\begin{figure}[htbp]
	\begin{center}
		\tikzstyle{state}=[shape=rectangle,rounded corners,draw=blue!50,fill=blue!20]
\tikzstyle{postate}=[shape=rectangle,rounded corners,draw=green!50,fill=green!20]
\tikzstyle{observation}=[shape=rectangle,rounded corners,draw=orange!50,fill=orange!20]
\tikzstyle{lightedge}=[<-,dotted]
\tikzstyle{mainstate}=[state,thick]
\tikzstyle{mainedge}=[<-,thick]

\begin{tikzpicture}[]

    \node          (time) at (4,9) {Time, $t\longrightarrow$};
    \node (hidden-1) at (1,8) {$h_t$};
	\node[state] (pop-model) at (4,8) {Population}
	edge [mainedge] (hidden-1);
	\node[observation] (rl-agent) at (4,6) {RL Agent};
	\node (hidden-2) at (7,8) {$h_{t+1}$};
    \node[postate] (sim-env) at (4,4) {Sim Env};
    \node (prev-action) at (0, 5) {$\{a_{t-1}, o_{t-1}, r_{t-1}\}$};

    \draw[->, thick] (pop-model) -- (hidden-2) node[midway, above] {$h_t^{\prime}$};
    \draw[->, thick] (pop-model) -- (rl-agent) node[midway, left] {$h_t^{\prime}$};
    \draw[->, thick] ([xshift=0.1cm]sim-env.north) -- ([xshift=0.1cm]rl-agent.south) node[midway, right] (env-return) {$o_t, r_t$};
    \draw[<-, thick] ([xshift=-0.1cm]sim-env.north) -- ([xshift=-0.1cm]rl-agent.south) node[midway, left] {$a_t$};

	\draw[->, thick] (env-return) to[out=0,in=180] (hidden-2);
    \draw[->, thick] (prev-action) to[out=0,in=180] (pop-model);
\end{tikzpicture}
	\end{center}
	\caption{A figure showing the time-dynamics of the interactions between the three components of \alg{}: The population model, the RL agent, and the Simulated Environment. }
	\label{fig:sim-edu}
\end{figure}
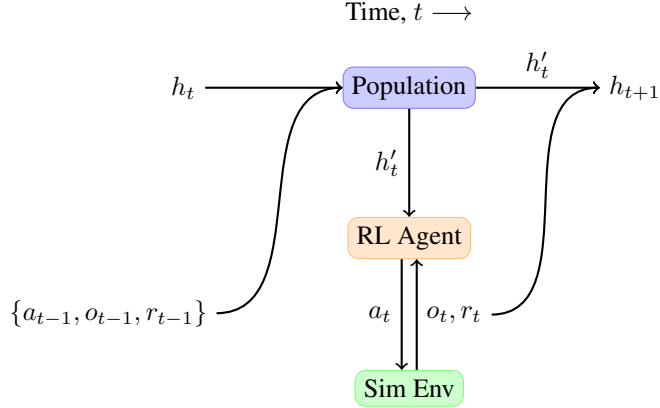

In order to address the difficulties of educational assistance, we use a reinforcement learning (RL) framework. RL allows us to address some of the issues of dynamically changing environments and population shifting, while still providing a computationally tractable approximation to dynamic programming searches. It also provides us with some basic algorithms to benchmark some of the ideas behind the education problem, highlight some of its difficulties, and understand what could be optimal approaches. \alg{} introduces a combination of 
\begin{enumerate}
    \item A highly configurable simulated education environment. In particular, we allow the configuration of concept structure, courses, student types, and possible interventions. Based on the simulated aspect, we also allow the exploration of the level of hidden information and how much information interventions can provide to the ITS.  
    \item A list of RL algorithms that demonstrate the use of the simulated education problem and provide solutions, both in a purely data-driven technique and with fixed heuristic-based models. The idea behind the variety of algorithms is for the demonstration of a point, whether it is for performance or explainability. 
    \item A knowledge tracing mechanism for population modeling. In the population modeling, there are the considerations of modeling student sub-types, using prior knowledge, and its sample efficiency with regards to individual modeling. 
\end{enumerate}
A view of the information flow is provided in Figure \ref{fig:sim-edu}.

In order to create the simulation, we draw from the wealth of teaching experience that is available. This allows us to simulate environments using rules-based structures, giving causal explanations for certain student responses. These experiences give us a prior understanding of how an educational course should function and inform our design decisions in the simulation process. 

Ultimately the goal of this simulated environment is to perform randomized experiments where our RL agent is interacting with a simulated human with predictable but stochastic tendencies. This does not necessarily have to be in the education envrionment, but education has some nice simplifications that we take advantage of in our problem that makes it easily parallel to a real-world situation. Ideally, up to some level of approximation, the simulated environment can directly reflect data that is gathered in the real world and, then, provide an ITS for future students. 

\subsection{Simulated Environment}

First, we introduce \alg{} as a highly-configurable, dynamic time-series environment used for RL probing. The simulated environment gives us a well-defined method of analyzing what conditions are necessary to provide good feedback in the ITS situation. 

\subsubsection{Concept Graphs}

Our first assumption about the education problem is that there exists a functional concept graph that guides the educational experience. For example, we normally know that some fundamental understanding of algebra is required before learning calculus, represented as a path in the graph Algebra $\rightarrow$ Calculus. In theory, we can propose a directed acyclic graph (DAG) of the learning concepts of an educational path. In our simulated environment, we primarily deal with linear connections between concepts. This embodies the assumption that there is no way to master a child concept before mastering a parent concept. 

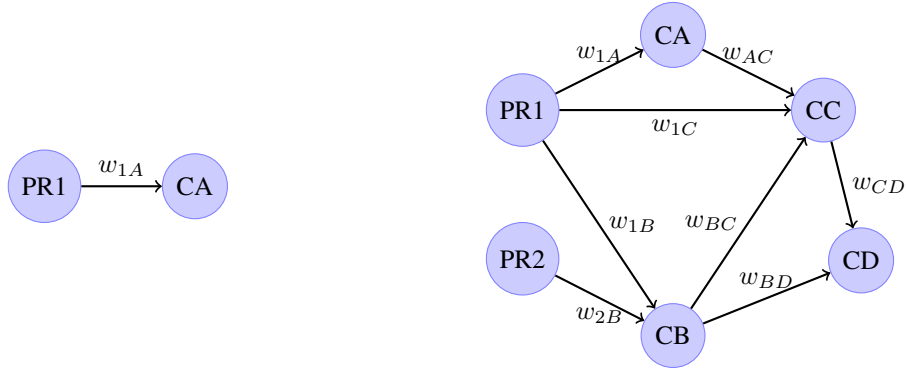
\begin{figure}
    \centering
    \begin{minipage}[c]{0.45\textwidth}
        \centering
        \begin{adjustbox}{valign=c}
            \usetikzlibrary{positioning}

\tikzstyle{state}=[shape=circle,draw=blue!50,fill=blue!20]
\tikzstyle{postate}=[shape=rectangle,draw=green!50,fill=green!20]
\tikzstyle{observation}=[shape=rectangle,draw=orange!50,fill=orange!20]
\tikzstyle{lightedge}=[<-,dotted]
\tikzstyle{mainstate}=[state,thick]
\tikzstyle{mainedge}=[<-,thick]

\begin{tikzpicture}[]
	\node[state] (prereq1) at (4.5,0) {PR1};
	\node[state] (conceptA) at (6.5, 0) {CA}
		edge [mainedge] node[above] {$w_{1A}$} (prereq1);

\end{tikzpicture}
        \end{adjustbox}
        \label{fig:prereq-one-concept}
    \end{minipage}
    \hfill
    \begin{minipage}[c]{0.45\textwidth}
        \centering
        \begin{adjustbox}{valign=c}
            \usetikzlibrary{positioning}

\tikzstyle{state}=[shape=circle,draw=blue!50,fill=blue!20]
\tikzstyle{postate}=[shape=rectangle,draw=green!50,fill=green!20]
\tikzstyle{observation}=[shape=rectangle,draw=orange!50,fill=orange!20]
\tikzstyle{lightedge}=[<-,dotted]
\tikzstyle{mainstate}=[state,thick]
\tikzstyle{mainedge}=[<-,thick]

\begin{tikzpicture}[]
	\node[state] (prereq1) at (4.5,0) {PR1};
	\node[state] (prereq2) [below=of prereq1] {PR2};
	\node[state] (conceptA) at (6.5, 1) {CA}
		edge [mainedge] node[above] {$w_{1A}$} (prereq1);
    \node[state] (conceptB) at (6.5, -3) {CB}
		edge [mainedge] node[right] {$w_{1B}$} (prereq1)
        edge [mainedge] node[below] {$w_{2B}$} (prereq2);
    \node[state] (conceptC) at (8.5, 0) {CC}
		edge [mainedge] node[below] {$w_{1C}$} (prereq1)
        edge [mainedge] node[above] {$w_{AC}$} (conceptA)
        edge [mainedge] node[left] {$w_{BC}$}(conceptB);
    \node[state] (conceptD) at (9, -2) {CD}
        edge [mainedge] node[above] {$w_{BD}$}(conceptB)
        edge [mainedge] node[right] {$w_{CD}$}(conceptC);
\end{tikzpicture}
        \end{adjustbox}
        \label{fig:four-concept}
    \end{minipage}
    \caption{Concept DAGs for multi-concept courses (left: one-concept course with prerequisite, right: four-concept course with two prerequisites). Concepts taught in a course are denoted with the precursor C and are denoted by letters (e.g. CA), while prerequisites are denoted with precursor PR and selected via numbers (e.g. PR1).}
    \label{fig:concept-graph}
\end{figure}

In practice, a full educational graph can get highly complex. Instead, we look at potential subgraphs, with the Markovian assumption that knowing the state of the parents of the subgraph, all future interactions are independent of the previous nodes. Two basic educational graphs that we use in our multi-concept courses can be seen in Figure \ref{fig:concept-graph}. 

Subgraphs also handle the different levels of granularity concept graphs can have. Within a course, small, individual concepts, \textit{e.g.} factoring, completing the square, and the quadratic formula as methods of solving quadratic equations, are taught one-by-one and may require specific attention. However, from the perspective of a follow-up course, an overall understanding of the higher-level concept is enough. Following the example, the three methods combine to form an understanding of how to solve a quadratic equation, which is what the next course needs. Even further along, this may even get further abstracted into ``algebra understanding.'' 

Any concept DAG will also have some weights $w_{\alpha\beta}$, representing how much the mastery of concept $\alpha$ affects concept $\beta$. Suppose we have concept $\gamma$. Let $p \in P(\gamma)$ represent the direct parents of $\gamma$. Then, the combined concept mastery $C_\gamma'$ can be represented as: 
\begin{equation} \label{eq:concept-parent}
    C_\gamma' = \sum_{p \in P(\gamma)} w_{p\gamma} C_p' + \left(1 - \sum_{p \in P(\gamma)} w_{p\gamma}\right) C_\gamma  
\end{equation}
where $1 - \sum_{p \in P(\gamma)} w_{p\gamma}$ represents the weight of the concept independent of any of its parents. This representation, therefore, requires a DAG structure, as it requires the finalized concept mastery for every parent $C_p'$ before $C_\gamma'$ can be computed. If $\gamma$ were ever an ancestor of a parent, Equation \ref{eq:concept-parent} would define a cyclic recursion.

\subsubsection{Students}
Once we have defined a concept graph, we can define students who have some inherent mastery of every relevant concept. The student also contains a couple of additional parameters: the amount of time they are willing to spend in each course time-step and a ``motivation'' parameter, a valued parameter ranging from 0 to 1 which affects the trajectory of student progress. In this simulation, there are two determining factors of the ``student type'': the initial state of the concepts (particularly relevant for prerequisite concepts) and their inherent motivation trajectory. This motivation parameter is an abstraction including their actual motivation, inherent study skills, discipline, and other external factors. 

As the motivation directly scales the effectiveness of the time spent, we do not also vary the amount of time for the student and include the variation of that abstraction within the motivation parameter. The time spent directly influences a small amount of immediate time reward regardless of its effectiveness. 

For our experiments, we test with several possible trajectories: stable trajectories starting at different levels and trajectories that trend upward and downward to represent students that need a warm-up period and students that burn out throughout a course's time. We also include some small noise to the motivation, based on the possibility of random external factors influencing student motivation. 


Students are generated using user-defined distribution specifications. In the current simulation, the specifications include priors about relevant concepts, primarily the prerequisite ones, and prior trajectories for motivation. In our current simulations, we assume that once these priors are set, they are fixed for the course, up to some random noise. 

\subsubsection{Interventions}
Once the students are defined, we can define possible interventions. An intervention is any interaction with the student that deals with the concepts or motivation of the student. There are \textit{scheduled interventions}, which are the scheduled lectures and examinations. We mark \textit{time-steps} as the time between scheduled interventions. Then, the \textit{actionable interventions} are the interventions that the ITS has access to that are dynamically chosen based on its observations. These interventions include probing interventions, tutor interventions, and motivation interventions. In our simulation, all actionable interventions have a positive cost associated with them, representing the time that each intervention takes. 

Our concept interventions, which constitutes both lectures and tutor interventions, are currently modeled as asymptotic exponential steps, given by 
\begin{equation}
    \frac{\Delta C}{\Delta t} = k_m(C_{target} - C)
\end{equation}
where $C$ represents the concept mastery, $k_m$ is some constant scaled by both the intervention type and the student's motivation, and $C_{target}$ can be defined per intervention type. 

Then, each concept intervention can also give some amount of feedback, where the configured quality of the feedback depends on the type of intervention. In our experiments, lectures provide no feedback, tutor sessions provide a good amount of feedback, and examinations provide a lot of feedback. In our simulation, we define the feedback as some number of i.i.d. Bernoulli samples ($\text{Bern}(C)$). 

Our motivation interventions, instead, are discrete steps. We divide the motivation into two main categories, a \textit{study skills} aspect and a \textit{transient motivation} aspect. We have a study-skills intervention that directly improves the study skills of a student permanently, but can only happen once. Then, we have a limited motivation intervention that can improve motivation for a student for a short period of time, in case there is a particular concept that the student needs the extra motivation for. Both interventions currently move the student up one step when active. 

Probing interventions provide little to no direct benefit to the student, but provides much more feedback. For our simulations, we offer two types of probes: a realistic probe and an oracle probe. The oracle probe directly elucidates the hidden parameter $C$ that defines the student's concept mastery. One can view this as learning the underlying probability distribution dictating the Bernoulli samples. This type of probe ameliorates the partial observability problem, as it allows the problem to become fully observed, at a cost. The realistic probe, instead, simulates an examination for low cost. This gives the ITS reliable information, but is subject to some random noise. Thus, it cannot give the same quality of information but it would be more realistic to implement. 

\subsubsection{Course}
A course is the list of \textit{scheduled interventions}. We design three types of courses, a \textit{basic one-concept} course, a \textit{prerequisite one-concept} course, and a \textit{four-concept} course. The prerequisite one-concept course and the four-concept course follow a concept graph structurally defined by Figure \ref{fig:concept-graph}. 

We design courses to have the following difficulty level: 
\begin{itemize}
    \item The top students (referred to as `A' students) will pass a course more than 90\% of the time with no ITS intervention. 
    \item The bottom students will almost never pass the course without ITS intervention. However, if the student receives significant ITS intervention, they should almost always pass the course. 
\end{itemize}
An ITS can always tutor to help students pass the course, but this would not respect all students' time. Therefore, the goal is to design an ITS that probes for the student type and then determines the amount of aid the student needs to receive.

\subsection{Population Model}
Following the definition of a course, another important aspect to an ITS's success is the population of students coming in. Hidden information is present everywhere in the education problem. We decide that there are primarily two main pieces of student information that will be hidden in the simulation, the student's concept mastery and the student's trajectory, which we abstract as the student's ``motivation.'' These two define the population of students that enter a course. 

The student concept mastery refers more to an instantaneous mastery, which can be noisily measured with an evaluation or examination of some kind. On the other hand, the student's motivation cannot be measured instantaneously and could depend on longer-term impacts and events. 

\subsubsection{Simulation}
Following the discussion of the simulated environment, the student population is defined with two points: the state of prerequisite concepts at the beginning of the course, and the overall student type in terms of motivation. Each prerequisite concept is quantized into 4 possible concept masteries, with most students in a natural classroom resting around the passing level of all prerequisite concepts, and 20\% students below and above that value. 

Everything about the distribution is hidden to the population model and the ITS, except perhaps in how many quantized steps we have split the distribution into. 

\subsubsection{Knowledge Tracing}
Knowledge tracing is the task of modeling the time-series nature of student knowledge in order to accurately predict student performance in the future, an inherently difficult problem due to the complexity of humans \cite{dkt}. We apply Bayesian inference in the form of a Hidden Markov Model (HMM) to provide knowledge tracing. However, we attach an additional component that gives us flexibility in explainability, use of prior knowledge, and exploration noise in reinforcement learning. 

When dealing with student concepts, we first discretize the concept understanding into $K = 4$ non-equal buckets. Using these buckets, we can model population information by maintaining initial, transition, and emission probability Dirichlet priors \cite{psrl, diggavi}. The Dirichlet distribution is the conjugate prior of the categorical distribution. This means that sampling from the Dirichlet distribution gives rise to possible categorical probability distributions required by the HMM. The Dirichlet distribution is also flexible, and different parameters can specify distributions with bell-shapes or long tails, depending on necessity. It is also specified with only as many paramters as the number of categories in the probability distributions. Finally, it has an intuitive Bayesian update, where observing a specific interaction can be updated by simply adding a weighted increment to that interaction. Thus, its parameters are always explainable as being the result of past observations and expert priors can be implemented easily.  

These give us very explainable ways to both use prior or expert knowledge and interpret the training as time goes on. We define Dirichlet parameters for each state-action pair. We use the parameters to sample initial, transition, and emission probability distributions in an epoch. However, as the emission probabilities depend only on how we want to define the categories of students, we do not need to learn the distributions. We can set them beforehand as a fixed prior to categorize student progress in each concept. The Dirichlet sampling also immediately provides exploration noise to the reinforcement learning agent, creating some input noise for robustness. From these distributions, we maintain estimated student concept masteries with respect to time, getting both a likelihood of being in each respective bucket, as well as the maximum likelihood state. As per the traditional HMM step \cite{rabiner89}, 
\begin{equation}
    \mathbb{P}\{ h_t = k \} = \frac{ \sum_{i = 1}^K \mathbb{P}\{ h_{t-1} = i \} P_T(i,k) P_E(k, o_t) }{\sum_{j=1}^K \sum_{i = 1}^K \mathbb{P}\{ h_{t-1} = i \} P_T(i,j) P_E(j, o_t) }
\end{equation}
Then, as we update our models, we also update the Dirichlet priors with the state transitions. 

Unfortunately, as the Dirichlet parameter method is still fundamentally mostly a tabular approach, this method can be significantly less sample efficient. Therefore, the selection of a relevant state space can heavily affect the training time and effectiveness of our approach in hidden education environments. For example, we have found that the inclusion of the actual value of the time step in the state is not desirable, and logically feels extraneous as doing a specific intervention should not have time-dependent properties.

\begin{figure}[htbp]
	\begin{center}
		\tikzstyle{state}=[shape=circle,draw=blue!50,fill=blue!20]
\tikzstyle{postate}=[shape=rectangle,draw=green!50,fill=green!20]
\tikzstyle{observation}=[shape=rectangle,draw=orange!50,fill=orange!20]
\tikzstyle{lightedge}=[<-,dotted]
\tikzstyle{mainstate}=[state,thick]
\tikzstyle{mainedge}=[<-,thick]
\begin{tikzpicture}[]
    \draw[thick,dotted,fill=green!20, opacity=0.4] (-1, 4.5)  rectangle (7,3.5);
    \node			   (student) at (3,4) {Population Model $\phi_T$};
    
    \node[state] (P1) at (0,0) {\small $h_{t-1}$};

    

    \node[state, minimum size=1cm] (ht) at (3, 0) {$h_t$}
	    edge [mainedge] node[above] (trans) {$P_T$} (P1);
        
    \node[observation, minimum size=0.75cm] (O1) at (3, 2.5) {$o_t$}
        edge [mainedge] node[left] (emis) {$P_E$} (ht);

    \node[state, minimum size=1cm] (htt) at (6, 0) {$\dots$}
        edge [mainedge] (ht);

    \node[postate] (rl) at (3, -2.5) {RL Agent}
        edge [mainedge] (ht);

    \draw let \p1 = (trans.north) in [->, thick] (\x1,3.5) -- (trans.north);
    \draw let \p1 = (trans.north) in [->, dashed] (\x1,3.5) to[out=270,in=180] (emis.west);
    \draw [->,dotted] (O1.north) -- (3,3.5);
    

\end{tikzpicture}

   
	\end{center}
	\caption{Bayesian knowledge tracing via a Dirichlet parameterized sampling technique. The population uses state information (which can possibly come from observations, such as the RL's choice of action) to sample $P_T \sim Dirichlet(\phi_T^{(o_t)})$. After sampling, we proceed with knowledge tracing with standard HMM updates. We can also sample a $P_E$ in a similar way, but in our case we use $P_E$ as fixed priors.}
	\label{fig:education-dag}
\end{figure}
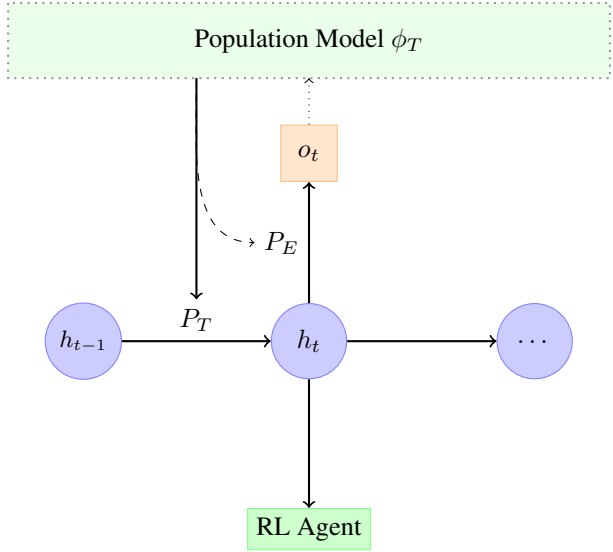

\subsubsection{Sub-Population Models}

In particular, one of the values in the state to consider is whether or not to consider sub-populations or student types. Because in our simulation we have defined the population in terms of prerequisite knowledge and motivation, this pair of values can determine a student type. Thus, there is an \textit{initial} student type and a \textit{transitional} student type, which dictates the motivation. In particular, by keeping student type in the model, we can have different Dirichlet parameters for each student type, creating a better understanding of possible student transitions, at the cost of reduced sample efficiency. 

However, trying to estimate student type is a second-order estimation problem, dependent on accurate estimation in the first order plus having priors in the second order. The estimation of the motivation parameter depends on an accurate estimation of the student concepts, but an accurate estimate of student concepts also depends on the estimation of the motivation parameter, leading to a somewhat cyclical estimation problem. This problem is \textit{intractable}, due to the accumulation of noise. Instead, we assume that students will always take a diagnostic quiz where the student self-assesses their student type. The population model can then use this information to select sub-populations. 

While we make the assumption that the population model has some prior information about which student type a student is in, this information is not passed to the RL agent, so any information the RL agent can gain can only be through a more accurate estimate of the student's state. In addition, even if the population model knows which group each student belongs in, it is not given direct information about what the group's expected motivation will be at every time step, or information about the dynamics of how motivation is integrated into the simulation. 

\subsubsection{Experiments}

We deal with environments with 3 levels of observability in our experiments. The first is a fully-observed environment, where we have all the oracle information necessary from the environment itself, so that if we allowed an oracle probe intervention for the agent, it would not actually do anything. Next, we deal with environments where only the concept information is hidden. This models an ITS that can ask a student what they believe their motivation is at the start of every session for free. However, the ITS cannot get accurate information about student concepts. Finally, we have an environment where both are hidden from the ITS. 

\subsection{Reinforcement Learning}

RL offers a general framework that specifically handles interactions with an unknown environment. As it focuses heavily on interactions, we look to RL as a way to be adaptive to different types of humans while interacting. 

However, RL still has many ongoing problems. In particular, unlike supervised learning, RL deals with a non-stationary non-independent dataset, making sample efficiency an important point of discussion. Therefore, RL has difficulty in low-interaction spaces and can have difficulty adjusting to a changing environment. 
Finally, RL still suffers from similar problems to supervised learning, in that interpretability and performance often have some level of tradeoff. 

\subsubsection{Rewards}
Overall, the goal that we have set for our ITS is to increase the passing rate of students, while giving students the most amount of free time. Thus, for time step $n$ in an $N$ step course, our reward function is defined as: 
\begin{equation} \label{eq:reward}
    R_n = \begin{cases}
        K_{g_n} \times g_n + K_\tau \times \frac{\tau_n}{T_n}& n < N \\
        K_{g_n} \times g_n + K_{pass} \times \mathbbm{1}_{G \ge G_{pass}} + K_\tau \times \frac{\tau_n}{T_n} & n = N
    \end{cases}
\end{equation}
where $K_{g_n}$, $K_\tau$, and $K_{pass}$ refer to the reward weights of the grade reward, time reward, and pass reward respectively. 
At every time step $n$, the environment provides a grade reward $g_n$ if a graded intervention takes place, and an immediate time reward, defined by the proportion of time the student has remaining after all the ITS interactions to how much she started with $\tau_n / T_n$. The immediate time reward can be realized as a student having free time to spend on other tasks or enjoyment, or, from the perspective of a tutor, the amount of time available for other students or tasks. At the end of the course we get the final course reward, based on the students' grade $G = \sum_{n} K_{g_n} g_n$, where we normalize with $\sum_n K_{g_n} + K_{pass} = 1$. For instance, for a one-concept course, we could have $K_{g_N} = 0.4$ (0 for all other values of $n$), $K_{pass} = 0.6$ and $G_{pass} = 75\%$.

\subsubsection{Rules-based Strategies}
As we use RL to learn some policies, we also want to compare the learned policies to some fixed rules-based policies. These rules-based policies come in several forms. First, we have the policies that do not respond to the student, which are the policy that never tutors and the policy that always tutors if time is available. These policies provide us with baselines for what we expect from non-interactive ITS that we can compare our RL agent and more interactive rules-based strategies with. 

Then, we add a layer of ``greedy'' interactivity. For each possible action of the ITS, we introduce a single interactive conditional for when each action should happen. We introduce a tutor limit. In this case, we would set the tutor limit to some value higher than $G_{pass}$, for instance 80\% or 85\%, such that we tutor until our estimate of the student's concept mastery exceeds this limit. We also introduce study-skills and nudge conditionals for when we perceive that the student's motivation is not maximized. Finally, we probe if the confidence of any concept is below a certain level. 

These rules-based strategies provide a stable greedy policy that we have confidence should be fairly successful to students. We call these strategies greedy because oftentimes the goal of the strategy is to simply take an action until it cannot be taken anymore or the condition for the action goes away. However, because of the simplicity of its design, it also provides highly explainable policies. We also note that the rules-based policies only make use of the same information as the model-free DQN and are not directly related to the rules used to design the simulation.



\section{Experiments}

We ran a series of comprehensive experiments involving the manipulation of various adjustable parameters, including motivation dynamics, external factor distributions, and the degree of partial observability. 

\subsection{Baseline Experiments}
First, we ran certain baseline experiments so that we could both confirm whether our courses were designed correctly and provide some understanding of the differences among heuristic policies. In these baselines, we typically assume a fully-observed environment. The fully-observed environment gives the ITS full detailed information about the underlying parameters used to compute the noisy observations. Furthermore, we can adjust the student populations to specifically check whether our course performs up to par. 

To reiterate, we expect that 
\begin{itemize}
    \item Using a no intervention policy, $A$ students should still pass the class most of the time, but $D$ students should almost never pass the course. 
    \item Using a tutor-only policy, almost every student should pass the course. However, as $A$ students should have passed the course anyway, they should receive a lower total reward. 
\end{itemize}

Results are shown in Table \ref{tab:baselines} and are largely in line with our expectations. ``All Students" describes a course with a student population that is within expectations, where most students are $B$ students with smaller percentages on either side. In all cases, without tutoring, $A$ students have a very high pass rate and a high test reward. After applying tutoring interventions, their pass rate goes up marginally, but their test reward always goes down. Similarly for $D$ students, their pass rate is close to 0 without any interventions, but reaches into 90\% when tutored. An interesting point is the definition of ``$D$ student'' produces few consistently poor students across as courses get more complex, as there are more prerequisite concepts, which can also factor into a student's success. Thus, even though in the four-concept courses, the $D$ students are still performing poorly when tutored every time, the absolute worst students are rare enough to not affect the total number of students much in a real distribution. 

\begin{table}[ht]
\centering
\begin{tabular}{@{}cclcc@{}}
\toprule
\begin{tabular}[c]{@{}c@{}}Course\\ Type\end{tabular} & Policy & \multicolumn{1}{c}{\begin{tabular}[c]{@{}c@{}}Student\\ Population\end{tabular}} & Test Reward & Pass Rate \\ \midrule
\multicolumn{1}{c|}{\multirow{6}{*}{\begin{tabular}[c]{@{}c@{}}Basic\\ One Concept\end{tabular}}} & \multirow{3}{*}{No Intervention} & \multicolumn{1}{l|}{All Students} & 1.0207 & \multicolumn{1}{r}{84.1\%} \\
\multicolumn{1}{c|}{} &  & \multicolumn{1}{l|}{$A$ Students} & 1.1171 & \multicolumn{1}{r}{97.9\%} \\
\multicolumn{1}{c|}{} &  & \multicolumn{1}{l|}{$D$ Students} & 0.4829 & \multicolumn{1}{r}{3.3\%} \\ \cmidrule(l){2-5} 
\multicolumn{1}{c|}{} & \multirow{3}{*}{Tutor Only} & \multicolumn{1}{l|}{All Students} & 1.0165 & \multicolumn{1}{r}{99.8\%} \\
\multicolumn{1}{c|}{} &  & \multicolumn{1}{l|}{$A$ Students} & 1.0245 & \multicolumn{1}{r}{100.0\%} \\
\multicolumn{1}{c|}{} &  & \multicolumn{1}{l|}{$D$ Students} & 0.9537 & \multicolumn{1}{r}{94.7\%} \\ \midrule
\multicolumn{1}{c|}{\multirow{6}{*}{\begin{tabular}[c]{@{}c@{}}Prerequisite\\ One Concept\end{tabular}}} & \multirow{3}{*}{No Intervention} & \multicolumn{1}{l|}{All Students} & 1.0228 & 93.2\% \\
\multicolumn{1}{c|}{} &  & \multicolumn{1}{l|}{$A$ Students} & 1.0832 & 100.0\% \\
\multicolumn{1}{c|}{} &  & \multicolumn{1}{l|}{$D$ Students} & 0.4208 & 4.8\% \\ \cmidrule(l){2-5} 
\multicolumn{1}{c|}{} & \multirow{3}{*}{Tutor Only} & \multicolumn{1}{l|}{All Students} & 1.0119 & 100.0\% \\
\multicolumn{1}{c|}{} &  & \multicolumn{1}{l|}{$A$ Students} & 1.0195 & 100.0\% \\
\multicolumn{1}{c|}{} &  & \multicolumn{1}{l|}{$D$ Students} & 0.9326 & 94.2\% \\ \midrule
\multicolumn{1}{c|}{\multirow{6}{*}{Four Concept}} & \multirow{3}{*}{No Intervention} & \multicolumn{1}{l|}{All Students} & 0.9632 & 63.3\% \\
\multicolumn{1}{c|}{} &  & \multicolumn{1}{l|}{$A$ Students} & 1.2057 & 99.7\% \\
\multicolumn{1}{c|}{} &  & \multicolumn{1}{l|}{$D$ Students} & 0.5267 & 0.0\% \\ \cmidrule(l){2-5} 
\multicolumn{1}{c|}{} & \multirow{3}{*}{Tutor Only} & \multicolumn{1}{l|}{All Students} & 1.0282 & 99.4\% \\
\multicolumn{1}{c|}{} &  & \multicolumn{1}{l|}{$A$ Students} & 1.0485 & 100.0\% \\
\multicolumn{1}{c|}{} &  & \multicolumn{1}{l|}{$D$ Students} & 0.6612 & 46.7\% \\ \bottomrule
\end{tabular}
\caption{A list of design baselines for the three courses that we use.}
\label{tab:baselines}
\end{table}




\subsection{Time Reward Experiments}


First, we experiment with the reward constants, primarily the time reward constant $K_\tau$. Intuitively, we can understand that as $K_\tau \rightarrow 0$, the reward depends only on the grade, and policies will tend toward tutoring as much as possible, as the cost for doing interventions goes to 0. On the flip side, as $K_\tau \rightarrow \infty$, it can totally dominate the grade aspect of the reward, and so the policy would no longer attempt to interact with the student whatsoever. 

Table \ref{tab:one-concept-time-rew} shows the adaptability of interactive policies in a fully-observed one-concept course. The first two columns serve as baseline comparisons, as they show the expectation as we move in either direction for $K_\tau$. As expected, the pass rates of the non-interactive strategies do not change across $K_\tau$ and only the test reward changes to reflect the effect of $K_\tau$. 

We observe that the DQN is able to adapt to the population differently based on $K_\tau$. For high levels of $K_\tau$, the DQN finds a policy with very high test reward, while finding a pass rate between the no intervention and tutor-only policies. One note is that there is some computational overhead present for extremely low values of $K_\tau$, as the reward does not propagate well into the DQN. 

Furthermore, we also include a \textit{tutor limit} rules-based approach approach, based primarily on our expectation of a good fully-observed policy. The tutor limit policy simply tutors until it reaches some limit, e.g. 85\% to be safe, and then stops tutoring. In all but the highest time reward, the rules-based approach actually performs similarly or better than the DQN in both fronts. This greedy method sacrifices some of the test reward it can get from the best students for the stability of making sure students always pass, but knows to stop at a certain point to retain a good amount of reward. These results illuminate one big difference between the RL algorithm approaches and the greedy rules-based approaches. The RL algorithms tend to be greedy toward the immediate rewards, procrastinating as much as possible and trying to get as much test reward through immediate rewards. On the other hand, the tutor limit policy does the exact opposite, where it is greedy toward the pass rate and tries for stability in that front. 

This implies that having a dynamic understanding of the reward can both be inherently useful and have a visible effect on students. Thus, this allows for the custom selection of how much time a student is willing to spend on a course and gives concrete results as to what the expected results would be for a student. Furthermore, we show that rules-based approaches provide both more stable and explainable solutions with similar results in most cases. 

\begin{table}[ht]
\centering
\scalebox{0.75}{
\begin{tabular}{rcccccccc}
\hline
\multicolumn{1}{c}{} & \multicolumn{8}{c}{Policy} \\ \hline
\multicolumn{1}{c|}{} & \multicolumn{2}{c|}{No Intervention} & \multicolumn{2}{c|}{Tutor Only} & \multicolumn{2}{c|}{DQN} & \multicolumn{2}{c}{Tutor Limit} \\ \hline
\multicolumn{1}{c|}{$K_\tau$} & Test Reward & \multicolumn{1}{c|}{Pass Rate} & \multicolumn{1}{c|}{Test Reward} & \multicolumn{1}{c|}{Pass Rate} & \multicolumn{1}{c|}{Test Reward} & \multicolumn{1}{c|}{Pass Rate} & Test Reward & Pass Rate \\ \hline
\multicolumn{1}{l|}{0.0001} & $0.800 \pm 0.253$ & \multicolumn{1}{c|}{80.8\%} & $0.959 \pm 0.026$ & \multicolumn{1}{c|}{99.9\%} & $0.940 \pm 0.010$ & \multicolumn{1}{c|}{98.6\%} & $0.941 \pm 0.044$ & 99.6\% \\
\multicolumn{1}{r|}{0.0005} & $0.811 \pm 0.247$ & \multicolumn{1}{c|}{81.8\%} & $0.959 \pm 0.025$ & \multicolumn{1}{c|}{99.9\%} & $0.943 \pm 0.015$ & \multicolumn{1}{c|}{98.6\%} & $0.944 \pm 0.044$ & 99.6\% \\
\multicolumn{1}{r|}{0.001} & $0.830 \pm 0.236$ & \multicolumn{1}{c|}{84.1\%} & $0.961 \pm 0.025$ & \multicolumn{1}{c|}{99.9\%} & $0.947 \pm 0.011$ & \multicolumn{1}{c|}{99.0\%} & $0.945 \pm 0.056$ & 99.3\% \\
\multicolumn{1}{r|}{0.002} & $0.830 \pm 0.244$ & \multicolumn{1}{c|}{82.6\%} & $0.964 \pm 0.026$ & \multicolumn{1}{c|}{99.9\%} & $0.949 \pm 0.009$ & \multicolumn{1}{c|}{98.4\%} & $0.954 \pm 0.033$ & 99.8\% \\
\multicolumn{1}{r|}{0.005} & $0.855 \pm 0.249$ & \multicolumn{1}{c|}{81.7\%} & $0.969 \pm 0.053$ & \multicolumn{1}{c|}{99.4\%} & $0.969 \pm 0.006$ & \multicolumn{1}{c|}{98.8\%} & $0.972 \pm 0.045$ & 99.6\% \\
\multicolumn{1}{r|}{0.01} & $0.901 \pm 0.252$ & \multicolumn{1}{c|}{81.1\%} & $0.984 \pm 0.054$ & \multicolumn{1}{c|}{99.4\%} & $0.999 \pm 0.007$ & \multicolumn{1}{c|}{98.4\%} & $1.003 \pm 0.054$ & 99.4\% \\
\multicolumn{1}{r|}{0.02} & $0.994 \pm 0.258$ & \multicolumn{1}{c|}{80\%} & $1.017 \pm 0.036$ & \multicolumn{1}{c|}{99.8\%} & $1.067 \pm 0.009$ & \multicolumn{1}{c|}{97.7\%} & $1.067 \pm 0.044$ & 99.7\% \\
\multicolumn{1}{r|}{0.05} & $1.309 \pm 0.245$ & \multicolumn{1}{c|}{82.5\%} & $1.106 \pm 0.061$ & \multicolumn{1}{c|}{99.6\%} & $1.3176 \pm 0.0197$ & \multicolumn{1}{c|}{95.3\%} & $1.255 \pm 0.075$ & 99.4\% \\ \hline
\end{tabular}}
\caption{Time Reward Experiments (One Concept)}
\label{tab:one-concept-time-rew}
\end{table}

One thing in particular shows a clear difference between the RL method and the rules-based method is when they tend to tutor. The rules-based approach is primarily greedily improving student concept mastery, up to a limit. On the other hand, the DQN (and many other RL algorithms) will tend to be greedy for time (gaining the instantaneous rewards).

\subsection{Partial Observability Experiments}

We apply different approaches to showcase the difficulty of partial observability and the effect of probing. Here, we use three different types of observability to make the problem more difficult. Concept-hidden environments hide the status of the concept mastery for the student from the ITS, but allow the ITS to see the motivation (based on some assumption that the student will tell the ITS at the beginning of each section). The unobserved environment hides both parameters. These are all done on the basic one concept setting with $K_\tau = 0.02$. 

We introduce the Study Skills (SS) heuristic strategies to indicate that the rules will attempt a study skills and/or nudge intervention when it deems that it would be helpful. The DQN class has access to these interventions, except potentially for the probe and the oracle probe intervention. In addition, when using any partially observed information environment, our policies will attach the relevant population model to track hidden information. Keep in mind that the tutor-only policy in this case does not need any population information here, as there is only a single concept to tutor. 

\begin{table}[ht]
\centering
\scalebox{0.8}{
\begin{tabular}{@{}crcccccc@{}}
\toprule
 & \multicolumn{1}{l}{} & \multicolumn{6}{c}{Policy Type} \\ \midrule
 & \multicolumn{1}{l|}{} & \multicolumn{2}{c|}{No Intervention} & \multicolumn{2}{c|}{Tutor Only} & \multicolumn{2}{c}{Probe Tutor Limit} \\ \midrule
Exp. Type & \multicolumn{1}{l|}{Population Type} & Test Reward & \multicolumn{1}{c|}{Pass Rate} & Test Reward & \multicolumn{1}{c|}{Pass Rate} & Test Reward & Pass Rate \\ \midrule
\multirow{4}{*}{Fully Observed} & \multicolumn{1}{r|}{Typical} & 1.0056 & \multicolumn{1}{c|}{81.9\%} & 1.0146 & \multicolumn{1}{c|}{99.5\%} & 1.0503 & 99.7\% \\
 & \multicolumn{1}{r|}{A Students} & 1.1170 & \multicolumn{1}{c|}{97.9\%} & 1.0226 & \multicolumn{1}{c|}{100.0\%} & 1.0721 & 100.0\% \\
 & \multicolumn{1}{r|}{D Students} & 0.4790 & \multicolumn{1}{c|}{2.9\%} & 0.9518 & \multicolumn{1}{c|}{94.2\%} & 0.9536 & 94.5\% \\
 & \multicolumn{1}{r|}{AD Students} & 0.7937 & \multicolumn{1}{c|}{49.8\%} & 0.9925 & \multicolumn{1}{c|}{97.9\%} & 1.0099 & 96.7\% \\ \midrule
\multirow{4}{*}{Concept Hidden} & \multicolumn{1}{r|}{Typical} & 0.9927 & \multicolumn{1}{c|}{79.8\%} & 1.0166 & \multicolumn{1}{c|}{99.8\%} & 1.0266 & 99.6\% \\
 & \multicolumn{1}{r|}{A Students} & 1.1151 & \multicolumn{1}{c|}{97.6\%} & 1.0237 & \multicolumn{1}{c|}{100.0\%} & 1.0451 & 100.0\% \\
 & \multicolumn{1}{r|}{D Students} & 0.4805 & \multicolumn{1}{c|}{3.1\%} & 0.9463 & \multicolumn{1}{c|}{93.5\%} & 0.9439 & 92.6\% \\
 & \multicolumn{1}{r|}{AD Students} & 0.7861 & \multicolumn{1}{c|}{48.4\%} & 0.9858 & \multicolumn{1}{c|}{96.9\%} & 0.9908 & 95.8\% \\ \midrule
\multirow{4}{*}{Unobserved} & \multicolumn{1}{r|}{Typical} & 1.0028 & \multicolumn{1}{c|}{81.4\%} & 1.0152 & \multicolumn{1}{c|}{99.7\%} & 1.0263 & 99.4\% \\
 & \multicolumn{1}{r|}{A Students} & 1.1174 & \multicolumn{1}{c|}{97.8\%} & 1.0255 & \multicolumn{1}{c|}{100.0\%} & 1.0454 & 100.0\% \\
 & \multicolumn{1}{r|}{D Students} & 0.4852 & \multicolumn{1}{c|}{3.8\%} & 0.9483 & \multicolumn{1}{c|}{93.6\%} & 0.9443 & 92.5\% \\
 & \multicolumn{1}{r|}{AD Students} & 0.7825 & \multicolumn{1}{c|}{47.9\%} & 0.9901 & \multicolumn{1}{c|}{97.4\%} & 0.9933 & 96.2\% \\ \midrule\midrule
\multicolumn{1}{l}{} & \multicolumn{1}{l}{} & \multicolumn{6}{c}{Policy Type} \\ \midrule
\multicolumn{1}{l}{} & \multicolumn{1}{l|}{} & \multicolumn{2}{c|}{SS Tutor} & \multicolumn{2}{c|}{Probe SS Tutor Limit} & \multicolumn{2}{c}{Oracle SS Tutor Limit} \\ \midrule
Exp. Type & \multicolumn{1}{l|}{Population Type} & Test Reward & \multicolumn{1}{c|}{Pass Rate} & Test Reward & \multicolumn{1}{c|}{Pass Rate} & Test Reward & Pass Rate \\ \midrule
\multirow{4}{*}{Fully Observed} & \multicolumn{1}{r|}{Typical} & 1.0168 & \multicolumn{1}{c|}{99.8\%} & 1.0657 & \multicolumn{1}{c|}{99.7\%} & 1.0535 & 100.0\% \\
 & \multicolumn{1}{r|}{A Students} & 1.0247 & \multicolumn{1}{c|}{100.0\%} & 1.0875 & \multicolumn{1}{c|}{100.0\%} & 1.0713 & 99.9\% \\
 & \multicolumn{1}{r|}{D Students} & 1.0058 & \multicolumn{1}{c|}{100.0\%} & 1.0312 & \multicolumn{1}{c|}{99.2\%} & 1.0169 & 99.7\% \\
 & \multicolumn{1}{r|}{AD Students} & 1.0152 & \multicolumn{1}{c|}{100.0\%} & 1.0557 & \multicolumn{1}{c|}{99.1\%} & 1.0414 & 99.7\% \\ \midrule
\multirow{4}{*}{Concept Hidden} & \multicolumn{1}{r|}{Typical} & 1.0188 & \multicolumn{1}{c|}{100.0\%} & 1.0614 & \multicolumn{1}{c|}{99.5\%} & 1.0282 & 99.9\% \\
 & \multicolumn{1}{r|}{A Students} & 1.0234 & \multicolumn{1}{c|}{100.0\%} & 1.0813 & \multicolumn{1}{c|}{99.9\%} & 1.0455 & 100.0\% \\
 & \multicolumn{1}{r|}{D Students} & 1.0062 & \multicolumn{1}{c|}{100.0\%} & 1.0089 & \multicolumn{1}{c|}{95.7\%} & 1.0090 & 99.6\% \\
 & \multicolumn{1}{r|}{AD Students} & 1.0144 & \multicolumn{1}{c|}{99.9\%} & 1.0472 & \multicolumn{1}{c|}{98.2\%} & 1.0264 & 99.8\% \\ \midrule
\multirow{4}{*}{Unobserved} & \multicolumn{1}{r|}{Typical} & 1.0158 & \multicolumn{1}{c|}{99.8\%} & 1.0585 & \multicolumn{1}{c|}{99.6\%} & 1.0052 & 99.9\% \\
 & \multicolumn{1}{r|}{A Students} & 1.0201 & \multicolumn{1}{c|}{100.0\%} & 1.0710 & \multicolumn{1}{c|}{100.0\%} & 1.0176 & 100.0\% \\
 & \multicolumn{1}{r|}{D Students} & 0.9979 & \multicolumn{1}{c|}{99.5\%} & 1.0016 & \multicolumn{1}{c|}{95.8\%} & 0.9677 & 97.5\% \\
 & \multicolumn{1}{r|}{AD Students} & 1.0085 & \multicolumn{1}{c|}{99.5\%} & 1.0355 & \multicolumn{1}{c|}{98.0\%} & 0.9962 & 99.3\% \\ \midrule\midrule
\multicolumn{1}{l}{} & \multicolumn{1}{l}{} & \multicolumn{6}{c}{Policy Type} \\ \midrule
\multicolumn{1}{l}{} & \multicolumn{1}{l|}{} & \multicolumn{2}{c|}{DQN No Probe} & \multicolumn{2}{c|}{DQN Probe} & \multicolumn{2}{c}{DQN All} \\ \midrule
Exp. Type & \multicolumn{1}{l|}{Population Type} & Test Reward & \multicolumn{1}{c|}{Pass Rate} & Test Reward & \multicolumn{1}{c|}{Pass Rate} & Test Reward & Pass Rate \\ \midrule
\multirow{2}{*}{Fully Observed} & \multicolumn{1}{r|}{Typical} & 1.0608 & \multicolumn{1}{c|}{92.1\%} & 1.0676 & \multicolumn{1}{c|}{93.2\%} & 0.9566 & 99.7\% \\
 & \multicolumn{1}{r|}{AD Students} & 1.0053 & \multicolumn{1}{c|}{96.8\%} & 1.0106 & \multicolumn{1}{c|}{99.9\%} & 1.0121 & 99.8\% \\ \midrule
\multirow{2}{*}{Concept Hidden} & \multicolumn{1}{r|}{Typical} & 1.0708 & \multicolumn{1}{c|}{93.8\%} & 1.0471 & \multicolumn{1}{c|}{98.1\%} & 1.0718 & 93.8\% \\
 & \multicolumn{1}{r|}{AD Students} & 1.0122 & \multicolumn{1}{c|}{100.0\%} & 0.9999 & \multicolumn{1}{c|}{97.9\%} & 1.0313 & 98.4\% \\ \midrule
\multirow{2}{*}{Unobserved} & \multicolumn{1}{r|}{Typical} & 1.0644 & \multicolumn{1}{c|}{96.4\%} & 1.0597 & \multicolumn{1}{c|}{92.1\%} & 1.0571 & 99.7\% \\
 & \multicolumn{1}{r|}{AD Students} & 1.0196 & \multicolumn{1}{c|}{99.5\%} & 0.9885 & \multicolumn{1}{c|}{97.2\%} & 1.0438 & 95.3\% \\ \bottomrule
\end{tabular}
}
\caption{Hidden Information Experiments Across a Variety of Policies. }
\label{tab:hidden}
\end{table}

Table \ref{tab:hidden} shows the results of multiple rules-based and DQN policies on these types of environments across differing population types. We include both low-entropy population types (single types of students) and high-entropy populations (an extreme $AD$ student distribution, which just has 50\% $A$ students and 50\% $D$ students) to get an idea for how much probing helps across the board. 

In the first row of Table \ref{tab:hidden}, we have the baseline effectiveness of the probe. The probe allows for accurate estimation with the population model, allowing for a limit in tutoring. There are two conclusions to be made. Across the board, the probe allows for better test reward, highlighting its effectiveness. However, the probes still do eat into tutoring time, showing some small pass rate drops for harder student distributions, yet test reward still improves in spite of the pass rate drop. 

Comparing the first and the second column shows the effectiveness of having study skills interventions compared to simply tutoring. Again, looking across the second column shows the effectiveness of probing. Interestingly, the oracle probe, even if at the same cost as a regular probe, will result in lower test reward, but higher pass rate, for the unobserved experiments. This is a reflection of its increased confidence in the student parameters, allowing it to more confidently reach the limit threshold of 85\%. However, this means that the oracle probe will have possibly used more tutors than necessary to account for the safety window. 

Then, we compare the second and third rows of Table \ref{tab:hidden}, as they have comparable intervention lists. The DQN policies are capable of finding policies that have similar in test reward to the heuristic methods, but usually with lower pass rates. However, with the same test reward, we typically prefer to have the higher pass rate. 

One final interesting result is that the DQN can produce highly variable results, especially with the addition of the probing and oracle interventions. The maximal result is presented in Table \ref{tab:hidden}, but a more in-depth analysis is presented in Appendix \ref{app:dqnedu}. Given all this, we see that there is no significant benefit of using RL in the complexities of a partially observed environment compared to our simple heuristic rules-based methods.

\subsection{Distributional Shift Experiments}
Here, we provide some insight into the robustness of the policies. We set up several experiments using the results we obtained in the previous experiment, specifically the population models and policies trained on the typical distribution and the one trained on the $AD$ student distribution in the completely unobserved environments. We test these models under both of the two original distributions, as well as one other distribution containing 25\% A students and 75\% D students. This $25/75$ distribution is a harder class in terms of necessary tutor interventions, so a lower test reward is expected, but it is slightly easier in terms of probing. Because the $AD$ population is closer to the $25/75$ distribution, we epxect that shift to be easier. 

Because we have both the population model and the policy as two independent components, we test situations where only a single one is mismatched. Table \ref{tab:dist-shift} shows the results of these distributional shifts. Immediately, the DQN shows to be more brittle to distributional change than the heuristic method. However, there is some nuance to the observation. 
\begin{enumerate}
    \item In cases where the population model is different from the test population, but the policy is the same, all policies perform only slightly worse than the in-distribution performance. This simulates a situation where a teacher may have some incorrect preconceived notions of the population of students, but follows a policy left behind by a previous teacher. This teacher performs almost similarly well to those who have trained heavily in that distribution. 
    \item On the other hand, when a teacher tries to force a policy trained on one distribution of students to a different one, things are not as stable. When going from a more difficult student distribution to an easier one (\textit{e.g.} from the $AD$ students to the typical student distribution), the teacher still performs well, but sacrifices some of the time required. On the other hand, when going from an easier student population to a harder one, the policy performs significantly worse than a trained model. These results are exacerbated with the $25/75$ distribution for the policy trained in a typical environment, but not for the one trained in the $50/50$ $AD$ environment. 
\end{enumerate}
These results show both the flexibility of the simulation and the heuristic models to distributional shifts in students. DQN shows some of the flexibility in these environments, but cannot perform well going from an easy student distribution to a harder one. 



\begin{table}[ht]
\centering
\scalebox{0.85}{
\begin{tabular}{@{}cccccccc@{}}
\toprule
\multicolumn{2}{c}{\textbf{SS Probe Tutor Limit}} & \multicolumn{6}{c}{Test Population} \\ \midrule
 & \multicolumn{1}{l|}{} & \multicolumn{2}{c|}{Typical Students} & \multicolumn{2}{c|}{$AD$ Students} & \multicolumn{2}{c}{$25/75$ $AD$ Students} \\ \midrule
\begin{tabular}[c]{@{}c@{}}Population\\ Model\end{tabular} & \multicolumn{1}{c|}{\begin{tabular}[c]{@{}c@{}}Policy\\ Distribution\end{tabular}} & Test Reward & \multicolumn{1}{c|}{Pass Rate} & Test Reward & \multicolumn{1}{c|}{Pass Rate} & Test Reward & Pass Rate \\ \midrule
Typical & \multicolumn{1}{c|}{Typical} & \textbf{1.0585} & \multicolumn{1}{c|}{\textbf{99.6\%}} & 1.0113 & \multicolumn{1}{c|}{99.6\%} & 0.9995 & 98.6\% \\
$AD$ Students & \multicolumn{1}{c|}{$AD$ Students} & 1.0211 & \multicolumn{1}{c|}{100.0\%} & \textbf{1.0355} & \multicolumn{1}{c|}{\textbf{98.0\%}} & 1.0056 & 99.4\% \\
Typical & \multicolumn{1}{c|}{$AD$ Students} & 1.0196 & \multicolumn{1}{c|}{99.8\%} & 1.0097 & \multicolumn{1}{c|}{99.3\%} & 1.0030 & 99.1\% \\
$AD$ Students & \multicolumn{1}{c|}{Typical} & 1.0203 & \multicolumn{1}{c|}{99.9\%} & 1.0106 & \multicolumn{1}{c|}{99.4\%} & 1.0056 & 99.3\% \\ \midrule
\multicolumn{2}{c}{\textbf{DQN}} & \multicolumn{6}{c}{Test Population} \\ \midrule
 & \multicolumn{1}{l|}{} & \multicolumn{2}{c|}{Typical Students} & \multicolumn{2}{c|}{$AD$ Students} & \multicolumn{2}{c}{$25/75$ $AD$ Students} \\ \midrule
\begin{tabular}[c]{@{}c@{}}Population\\ Model\end{tabular} & \multicolumn{1}{c|}{\begin{tabular}[c]{@{}c@{}}Policy\\ Distribution\end{tabular}} & Test Reward & \multicolumn{1}{c|}{Pass Rate} & Test Reward & \multicolumn{1}{c|}{Pass Rate} & Test Reward & Pass Rate \\ \midrule
Typical & \multicolumn{1}{c|}{Typical} & \textbf{1.0644} & \multicolumn{1}{c|}{\textbf{96.4\%}} & 0.8988 & \multicolumn{1}{c|}{71.4\%} & 0.8039 & 58.2\% \\
$AD$ Students & \multicolumn{1}{c|}{$AD$ Students} & 1.0278 & \multicolumn{1}{c|}{99.9\%} & \textbf{1.0196} & \multicolumn{1}{c|}{\textbf{99.5\%}} & 1.0101 & 99.0\% \\
Typical & \multicolumn{1}{c|}{$AD$ Students} & 1.0286 & \multicolumn{1}{c|}{99.9\%} & 1.0161 & \multicolumn{1}{c|}{99.0\%} & 1.0092 & 98.7\% \\
$AD$ Students & \multicolumn{1}{c|}{Typical} & 1.0649 & \multicolumn{1}{c|}{96.3\%} & 0.8918 & \multicolumn{1}{c|}{70.4\%} & 0.7940 & 56.7\% \\ \bottomrule
\end{tabular}
}
\caption{Distributional Shift Experiments. The table shows test reward results and pass rate of students when changing the population model, the policy, or both. Bolded values are in-distribution results. }
\label{tab:dist-shift}
\end{table}

\subsection{Structure Experiments}
\begin{figure}[htbp]
	\begin{center}
		\tikzstyle{state}=[shape=circle,draw=blue!50,fill=blue!20]
\tikzstyle{postate}=[shape=rectangle,draw=green!50,fill=green!20]
\tikzstyle{observation}=[shape=rectangle,draw=orange!50,fill=orange!20]
\tikzstyle{lightedge}=[<-,dotted]
\tikzstyle{mainstate}=[state,thick]
\tikzstyle{mainedge}=[<-,thick]
\begin{tikzpicture}[]
    \draw[thick,dotted,fill=green!20, opacity=0.4] (-1.5, 9.5)  rectangle (10.5,8.5);
    \node			   (student) at (4.8,9) {Student State};
    \node          (pg) at (-1.5,8) {Prerequisites};
    \node              (time) at (3,10) {Time, $t\longrightarrow$};
    
    \node[state] (P0) at (-1.5,6) {\small $PR_1$};
    \node[state] (P1) at (-1.5,2) {\small $PR_2$};

    \node[state] (C0) at (1, 7) {\small $CA$}
    	edge [mainedge] (P0);
    \node[state] (C1) at (3.5, 2) {\small $CB$}
        edge [mainedge] (P0)
	    edge [mainedge] (P1);
    \node[state] (C2) at (6, 7) {\small $CC$}
    	edge [mainedge] (C0)
    	edge [mainedge] (C1);
    \node[state] (C3) at (8.5, 4) {\small $CD$}
    	edge [mainedge] (C2)
    	edge [mainedge] (C1);

   \draw[->, thick] (P0) to[out=0,in=225] (C2);
    	
    \draw let \p1 = (C0.north) in [->, dashed] (\x1,8.5) -- (C0.north);
    \draw[->, dashed] (3.5,8.5) -- (C1.north);
    \draw[->, dashed] (6,8.5) -- (C2.north);
    \draw let \p1 = (C3.north) in [->, dashed] (\x1,8.5) -- (C3.north);

   	\draw[thick,dotted,fill=red!20, opacity=0.4] (1.75,1)  rectangle (2.75,7.5);
   	\node              (I0) at (2.25, 0.5) {$Q_1$};
   	\draw[thick,dotted,fill=red!20, opacity=0.4] (4.25,1)  rectangle (5.25,7.5);
   	\node              (I1) at (4.75, 0.5) {$M$};
   	\draw[thick,dotted,fill=red!20, opacity=0.4] (6.75,1)  rectangle (7.75,7.5);
   	\node              (I2) at (7.25, 0.5) {$Q_2$};
    \draw[thick,dotted,fill=red!20, opacity=0.4] (9.25,1)  rectangle (10.25,7.5);
   	\node              (I3) at (9.75, 0.5) {$F$};
   	
   	\draw let \p1 = (I3.south east), \p2 = (P1.south east) in [thick,dotted] ($(P0.north west)+(-0.5,1.45)$) rectangle ($(\x2, \y1)+(0.5, 0)$);
   	\draw[thick,dotted] ($(P0.north east)+(1,1.45)$)  rectangle ($(I3.south east)+(0.5, 0)$);
   	\node              (pg) at (4.8,8) {Current Course};

\end{tikzpicture}
	\end{center}
	\caption{The time-dynamic DAG structure of the four-concept course with the concept DAG referenced in \ref{fig:concept-graph}. The structure experiments are defined based on which subset of evaluations ($Q_1$, $M$, $Q_2$, and $F$) are present. }
	\label{fig:structure-exp}
\end{figure}
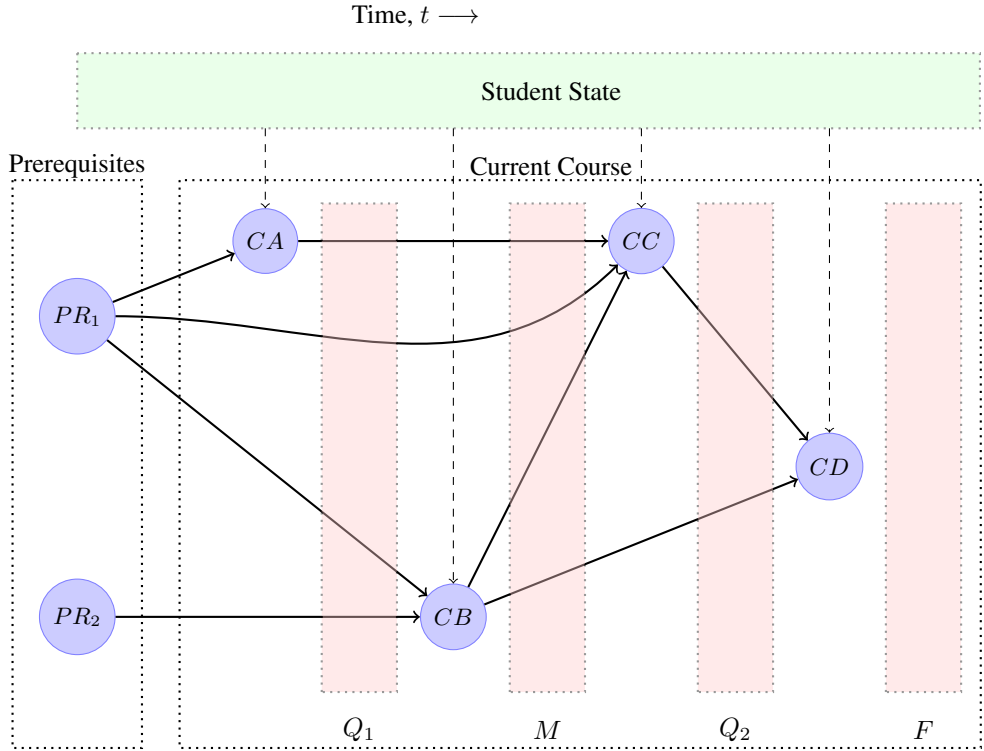

Finally, based on our observations on the effect of probing interventions, we suspect that we can also encode improved probing directly into the structure of courses. 
For the purposes of the experiment, we design courses assuming the ITS has limited probing capabilities, as many courses are designed today. 

Referencing Figure \ref{fig:structure-exp}, we create 4 different course structures. The first has only finals, testing all 6 concepts in the four-concept course at the end $F$. The second has a midterm-final structure, testing PR1, PR2 and CA at $M$ and CB, CC, and CD at $F$. Finally, we have a quiz structure that uses all 4 examinations $Q_1$, $M$, $Q_2$, and $F$. The final one is an extension of the quiz structure with an additional diagnostic quiz about halfway into the learning of every concept (and at the beginning of review sections for pre-requisite material). The understanding is that with more examinations, the ITS and instructor will have more information about the student, to become capable of adjusting their understanding of the students for more individualized plans. However, the more frequent examinations increases the urgency of tutoring, as the earlier grades can affect the overall total grade in the course. In order to fully align our goals with the balance of probing, we use the $AD$ distribution to maximize the difference between student types. 

Table \ref{tab:structure} shows results with these four different structures, tested on a random policy, a study-skills tutor-limit policy, and a DQN policy, each without the ability to probe. The random policy effectively shows the difficulty of the class, showing that in our simulation, the difficulty of the class increases with the complexity of the course structure. This is somewhat realistic, in that students who are time-limited lose some flexibility in when they can study for their course. However, this is also a limitation of the design of our system. Students do not forget in this environment, so students are never better off on an exam being tested earlier rather than later. 

Moving on to the heursitic and the learned policies, we see that they both achieve much higher test rewards and pass rates compared to the random policy. Overall, with the four concept policy, we see that the DQN's test reward is higher than that of the heuristic policy, but with a lower pass rate. Most importantly, both policies are able to pull the pass rate of the quiz and midterm-final structures up in line with the finals-only structures. While the policies are unable to allow them to surpass the finals-only structure, the amount of improvement showcases the utility of the extra information available in other structures in improving performance. 
Furthermore, the diagnostics provided in the diagnostic structure does provide a small boost in both methods.

\begin{table}[ht]
\centering
\begin{tabular}{@{}l|cc|cc|cc@{}}
\toprule
 & \multicolumn{2}{c|}{Random} & \multicolumn{2}{c|}{SS Tutor Limit} & \multicolumn{2}{c}{DQN} \\ \midrule
Course Structure & Test Reward & Pass Rate & Test Reward & Pass Rate & Test Reward & Pass Rate \\ \midrule
Finals Only & 1.0101 & 92.4\% & 1.0235 & 99.7\% & 1.0782 & 97.4\% \\
Midterm-Final & 0.9732 & 86.9\% & 1.0171 & 99.1\% & 1.0575 & 96.5\% \\
Quizzes & 0.9444 & 82.6\% & 1.0145 & 98.8\% & 1.0453 & 97.8\% \\
Quiz + Diags & 0.9596 & 85.0\% & 1.0164 & 98.8\% & 1.0659 & 94.7\% \\ \bottomrule
\end{tabular}
\caption{Four-Concept Structure Experiments. }
\label{tab:structure}
\end{table}

\section{Conclusion}
In this chapter, we develop a new simulation environment to explore adaptive assistance of student learning. The system captures the hidden information of learning, the tradeoffs in different types of interventions, and the value of probing in the context of a course. We find that for such systems, a deep RL agent performs well, but not significantly better than certain rules-based heuristic systems. It is shown that partial observability directly correlates to problem difficulty and thus, highlights the importance of probing. With the importance of probing, we also show that course structures that encourage more information gain show improvements in student performance. In general, we find that RL does not perform significantly better than well-designed heuristic methods in these hidden information environments, especially in terms of flexibility and explainability. 

There are still many ways to extend the work presented in this chapter. There are several extensions just to this simulated environment. Many of the design decisions in the simulation, though complex already, are made to simplify estimation as much as possible. First, one can extend the population model to have different levels of representation, explainability, and trainability. For instance, applying \cite{dkt} could provide a much improved population model with much less explainability. We can add some more overarching dynamics, such as forgetting, concept reinforcement via contextual reuse, and concept improvement dependencies across prerequisites. 

While the simulation has broad coverage over many hypothetical scenarios, we can use pre-existing observational data to make the simulation more realistic. Further mathematical models can be explored that more closely align with students' true concept graphs, trajectories, and conditional dependencies of traits. Plus, education is only one of the many time-series domains that are increasingly aided with computers. Many of the ideas here, though often with different names, can extend to other domains, such as medicine, psychology, personal assistance, and other areas of computer interactions.

\bibliographystyle{IEEEtran}
\bibliography{simedu}
\newpage
\onecolumn
\appendix
\section{Appendix}

\subsection{DQN and SimEdu} \label{app:dqnedu}

\begin{figure} [htb]
    \centering
    \includegraphics[width=\textwidth]{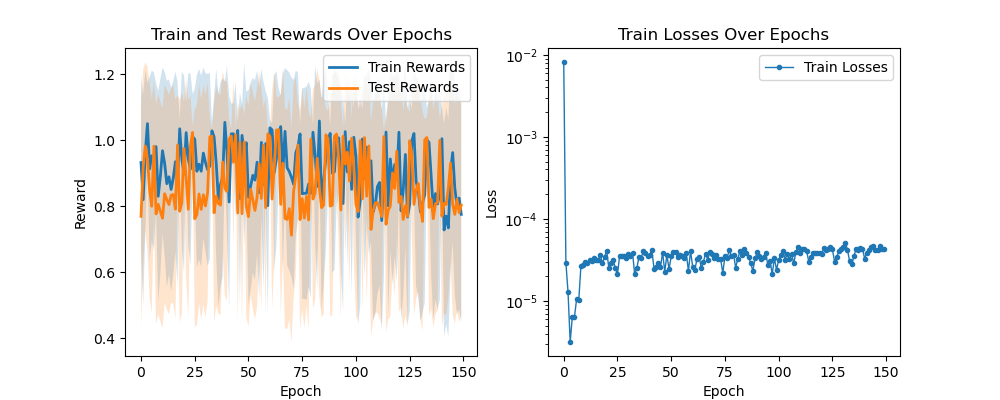}
    \caption{Example DQN Training Trajectory for DQN without probing capabilities on an unobserved course. Error bars represent the one standard deviation away from the mean across 1000 simulated students. }
    \label{fig:dqn-training-traj}
\end{figure}
We observed that DQN, when used with SimEdu, produces a very unstable and inconsistent training process. An example training trajectory is shown in Figure \ref{fig:dqn-training-traj}. Notice that within 25 epochs, the training loss has stabilized for the most part. However, the train and test rewards are highly inconsistent. Based on these results, we devise an evaluation metric to choose the best RL agent across the epochs. This evaluation metric is a weighted average between the average test reward, the median test reward, and the pass rate. 

Part of the difficulty comes from the design of the reward function, Equation \ref{eq:reward}. Because of the $\mathbbm{1}_{G \ge G_{pass}}$, there is a steep drop in reward when the total grade is low. Therefore the variance of rewards can be very large in a specific epoch. 

\begin{figure} [htb]
    \centering
    \includegraphics[width=\textwidth]{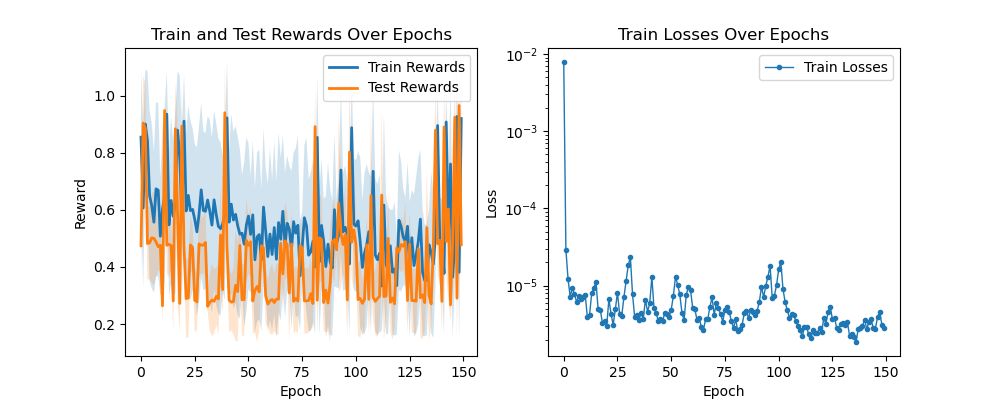}
    \caption{Example DQN Training Trajectory for DQN with full probing capabilities on an unobserved course. Error bars represent the one standard deviation away from the mean across 1000 simulated students. }
    \label{fig:dqn-training-traj-2}
\end{figure}

However, most of the inconsistency is likely due to the difficulty of partial observability for RL. In particular, we noticed that with a DQN that is capable of (oracle) probing, such as the one in Figure \ref{fig:dqn-training-traj-2}, the inconsistencies are even greater, and they tend to perform worse, even though they have actions that should help it within the dynamics. 
In the implementation, the DQN without probing capabilities has a reduced action space, which overall reduces the required search space. The low immediate impact of individual probes introduces several problems. First, introducing probes elongates the total number of steps while training, diluting rewards even further backward and requiring additional searches. Second, because probes can change the state so little, they can produce loop structures in the RL search space, causing possible confusion and continuous probing. Overall, the intuitive thought where RL is greedier in terms of immediate rewards shines through with the variability of results. We acknowledge that further hyperparameter tuning is required and the RL agents with the larger action space likely require longer training time. 

\end{document}